\documentclass[conference]{IEEEtran}
\IEEEoverridecommandlockouts
\usepackage{cite}
\usepackage{amsmath,amssymb,amsfonts}
\usepackage{algorithmic}
\usepackage{graphicx}
\usepackage{url}
\usepackage{booktabs}
\usepackage{textcomp}
\usepackage{xcolor}
\def\BibTeX{{\rm B\kern-.05em{\sc i\kern-.025em b}\kern-.08em
    T\kern-.1667em\lower.7ex\hbox{E}\kern-.125emX}}
\begin{document}

\title{Ordinal Encoding as a Regularizer in Binary Loss for Solar Flare Prediction}

\author{
\IEEEauthorblockN{
Chetraj Pandey\IEEEauthorrefmark{1}\IEEEauthorrefmark{4}, 
Jinsu Hong\IEEEauthorrefmark{2}, 
Anli Ji\IEEEauthorrefmark{3}, 
Rafal A. Angryk\IEEEauthorrefmark{2}, 
Berkay Aydin\IEEEauthorrefmark{2}}
\IEEEauthorblockA{\IEEEauthorrefmark{1}Department of Computer Science, Texas Christian University, Fort Worth, TX, USA}
\IEEEauthorblockA{\IEEEauthorrefmark{2}Department of Computer Science, Georgia State University, Atlanta, GA, USA}
\IEEEauthorblockA{\IEEEauthorrefmark{3}Department of Computer Science, California State University, Fullerton, CA, USA}
\IEEEauthorblockA{\IEEEauthorrefmark{4}Corresponding Author: c.pandey@tcu.edu}
}

\maketitle

\begin{abstract}
The prediction of solar flares is typically formulated as a binary classification task, distinguishing events as either Flare (FL) or No-Flare (NF) according to a specified threshold (e.g., $\geq$C-class, $\geq$M-class, or $\geq$X-class). However, this binary framework neglects the inherent ordinal relationships among the sub-classes contained within each category (FL and NF). Several studies on solar flare prediction have empirically shown that the most frequent misclassifications occur near this prediction threshold. This suggests that the models struggle to differentiate events that are similar in intensity but fall on opposite sides of the binary threshold. To mitigate this limitation, we propose a modified loss function that integrates the ordinal information among the sub-classes of the binarized flare labels into the conventional binary cross-entropy (BCE) loss. This approach serves as an ordinality-aware, data-driven regularization method that penalizes the incorrect predictions of flare events in close proximity to the prediction threshold more heavily than those away from the boundary during model optimization. By incorporating ordinal weighting into the loss function, we aim to enhance the model’s learning process by leveraging the ordinal characteristics of the data, thereby improving its overall performance.
\end{abstract}


\section{Introduction} \label{sec:intro}
Natural hazards such as earthquakes, tornadoes, volcanic eruptions, and space weather events are often described using severity indices that quantify their potential impact. These indices may be based on linear scales, such as flood severity \cite{flood} or tornado classifications \cite{tornadoes}, or logarithmic scales, such as those used for earthquakes \cite{earthquake}, volcanic eruptions \cite{volcanoes}, and solar phenomena like flares and energetic particle events \cite{spaceweather}. Predictive models for such events typically rely on fixed thresholds to separate classes for classification settings. However, these approaches often ignore the ordinal structure inherent in severity indices, which can be leveraged to improve modeling. In our previous work \cite{dsaa2024}, focused on solar flare prediction, we proposed a loss weighting strategy that penalized misclassifications of extreme flare instances, such as strong flares (X \& M) and flare-quiet (FQ) events, more heavily than those involving flare sub-classes in the middle of the scale. The assumption was that these errors are more critical and less acceptable. However, most misclassifications in practice occur near class boundaries, particularly between C- and M-class flares \cite{PandeySimBig2022, pandeyaike2023}. In this paper, within the context of solar flare prediction, we revisit that design choice and introduce a proximity-based penalty that emphasizes borderline errors, with the goal of reducing false positives and false negatives in regions where adjacent classes are most frequently confused by the model.

Solar flares are brief but intense releases of energy that occur on the Sun’s surface, emitting large amounts of extreme ultraviolet and X-ray radiation. They constitute a major area of interest in space weather forecasting. The National Oceanic and Atmospheric Administration (NOAA) classifies flares by their peak X-ray flux into five categories: X $(>10^{-4},Wm^{-2})$, M $(>10^{-5},Wm^{-2})$, C $(>10^{-6},Wm^{-2})$, B $(>10^{-7},Wm^{-2})$, and A $(>10^{-8},Wm^{-2})$ \cite{spaceweather}. These categories follow a logarithmic scale, where intensity decreases from X to A. Flares weaker than the A-class threshold are generally undetectable and are termed flare-quiet (FQ) \cite{pandeyaike2023, pandeyecml2024, dsaa2024}. Although M- and X-class flares occur infrequently, they are far more energetic than other flare classes and are of particular concern because of their potential to disturb near-Earth environment and disrupt technologies such as satellites, GPS, power grids, and aviation systems \cite{Yasyukevich2018}. Accordingly, binary solar flare prediction is often defined as forecasting the occurrence of flares with intensities at or above the M-class threshold.

In solar flare forecasting, binary classification is generally employed to distinguish flares according to their intensity levels. When the threshold is defined as $\geq$M, M- and X-class flares are categorized as Flare (FL), whereas C-, B-, and A-class flares, along with flare-quiet (FQ) instances, are grouped as No Strong Flare (NF). This binary formulation simplifies the task by separating high-impact events from less intense or negligible activity, which is useful for assessing potential solar disruptions. However, this approach ignores the ordinal relationships between flare sub-classes during model training. Standard loss functions such as cross-entropy and focal loss~\cite{focal} can be weighted to address class imbalance; however, they inherently treat class labels as nominal and therefore fail to capture the ordinal relationships among sub-classes within the FL and NF categories~\cite{dsaa2024}. These losses penalize all misclassifications equally, without considering the difference in severity between, for example, misclassifying a C-class and an A-class flare.

Numerous studies have examined different approaches to predicting solar flares, including human-based forecasting techniques \cite{Crown2012}, statistical models \cite{Lee2012}, and numerical simulations grounded in physical modeling \cite{Kusano2020}. In recent years, data-driven methods based on machine learning and deep learning have received increased attention due to their ability to process large datasets and their promising results in space weather forecasting \cite{Whitman2022}. There are several studies \cite{Bobra2015, Nishizuka_2017, Nishizuka2018, Huang2018, Leka2018, Li2020, Ji2020, pandey2021bigdata, Nishizuka2021, PandeySimBig2022, Pandey2022f, hong23, PandeyECML2023, Ji2023, hond3rd, pandeyaike2023, hong2nd, pandeyecml2024} that treat solar flare prediction as a binary forecasting problem, employing binary loss functions without incorporating any ordinal characteristics of flare classes\footnote{It is important to note that the literature exhibits substantial variability in data modalities, encompassing inputs from diverse sources and instruments. Differences are also evident in prediction targets (e.g., $\geq$C-, $\geq$M-, and $\geq$X-class flares), spatial resolution (full-disk versus active region-based data), data partitioning strategies, and forecasting horizons (e.g., 24 and 48 hours). Consequently, the performance of our models is not directly compared with studies employing such differing configurations, as they are not strictly comparable to the present work.}. Some of the studies (e.g., \cite{PandeySimBig2022, PandeyDS2023, pandeyaike2023, PandeyECML2023}) have shown that the most incorrect predictions of flares in a binary setting ($\geq$M) occur with borderline flare classes (i.e., C- and M-class flares near the classification boundary). This highlights the model's inability to distinguish subtle differences between flares in proximity of the threshold, leading to misclassification around the decision boundary. Introducing different costs for incorrect predictions, especially for borderline cases, can help the model minimize more costly errors. 

Therefore, in this study, we incorporate the ordinal characteristics of flare classes into the binary cross-entropy (BCE) loss through the use of weighting factors building upon our prior work in \cite{dsaa2024}. This method assigns class-dependent weights to instances according to their flare sub-class, enabling the model to capture the subtle distinctions that exist within each binary category (NF and FL). Specifically, we introduce a threshold proximity penalty (PP) into the BCE loss based on the ordinal nature of solar flares such that incorrect predictions for instances near the binary classification boundary are penalized more heavily, as these are the instances the model struggles to predict correctly compared to instances further from the chosen threshold. We hypothesize that this proposed adjustment of data-driven regularization to BCE loss, considering the ordinal nature of flare events will guide training of models towards better optimization and improve the models' performance.

The remainder of this paper is organized as follows. Section~\ref{sec:data} outlines the dataset used in this work and the overall architecture of the proposed flare prediction model. Section~\ref{sec:method} details the modification introduced to the standard cross-entropy loss and explains its application in the context of solar flare forecasting. Section~\ref{sec:expt} describes the experimental setup, including hyperparameter settings and evaluation results that demonstrate the effectiveness of the proposed approach. Finally, Section~\ref{sec:conc} concludes the paper by summarizing the main findings and discussing potential directions for future research.

\section{Data and Model} \label{sec:data}

\begin{figure}[tbp!]
\centering
\includegraphics[width=\linewidth ]{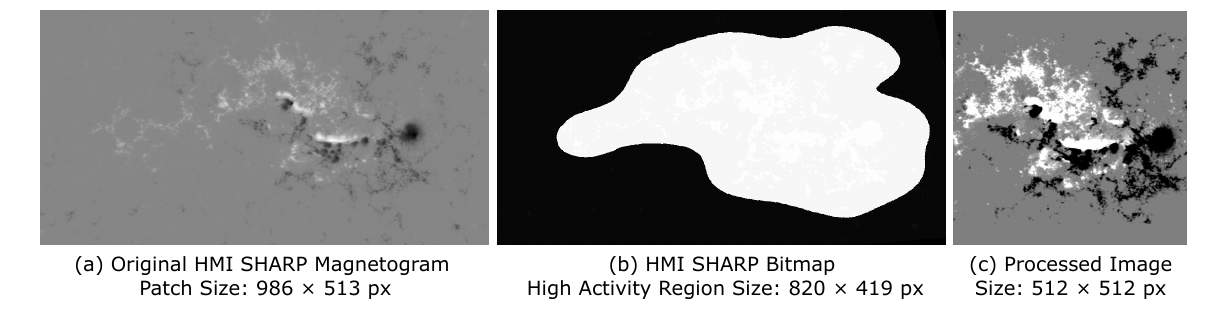}
\caption[]{An illustrative example showing: (a) the original raw HMI active region (AR) magnetogram corresponding to HARP~4781 at timestamp 2014-11-08T00:00:00~UTC; (b) the associated bitmap of the AR patch in (a), where white pixels indicate the region of interest; and (c) the final preprocessed AR image from (a), cropped to 512$\times$512, which is used for model training.}
\label{fig:dataexample}
\end{figure}
The dataset used in this research \cite{dataset} is a preprocessed version of the publicly available Spaceweather HMI Active Region Patches (SHARP) data product \cite{Bobra2014}, derived from line-of-sight (LOS) magnetograms of active regions (ARs) provided by the Helioseismic and Magnetic Imager (HMI) \cite{Schou2011} onboard the Solar Dynamics Observatory (SDO) \cite{Pesnell2011}. The data is preprocessed using the pipeline described in \cite{pandeyecml2024}. Specifically, we utilize hourly sampled LOS magnetograms of ARs located within $\pm$60$^\circ$ of the solar longitude from 2010 to 2018. The raw magnetograms contain high-resolution raster values of magnetic field strength, typically ranging from approximately $\pm$4500 G. Flux values are clipped at $\pm$256 G, and those within $\pm$25 G are set to zero to reduce noise. Additionally, bitmaps from the SHARP series are used to isolate the region of interest (ROI) within each AR. Bitmap-cropped patches smaller than 512$\times$512 pixels are zero-padded, while larger patches are downsampled by selecting a 512$\times$512 region containing the highest total unsigned flux (USFLUX). 

This selection strategy helps preserve spatial regions most relevant to flare activity. All patches are finally scaled to the range 0–255 to generate standardized image representations. Examples of raw magnetograms, extracted ROIs, and processed input images are shown in Fig.~\ref{fig:dataexample}(a–c). For each active region (AR) patch, a binary label is assigned based on the peak X-ray flux as follows: (i) $\geq$M corresponds to a Flare (FL) event, indicating the occurrence of relatively strong flaring activity, and (ii) $<$M corresponds to No Strong Flare (NF), defined within a 24-hour prediction window. Specifically, if the maximum NOAA flare class observed within 24 hours from the timestamp of a given AR patch is $<$M, the sample is labeled as NF; otherwise, it is labeled as FL.

\begin{table}[tbh!]
\setlength{\tabcolsep}{5pt}
\renewcommand{\arraystretch}{1.5}
\caption{The distribution of flare classes in relation to binary classes, showing training and validation and test set.}
\vspace{-10pt}
\begin{center}
\begin{tabular}{rrrrrr}
\toprule
\textbf{Binary} & \textbf{Flare} & \textbf{Train Set} & \textbf{Train Set} & \textbf{Val.} & \textbf{Test} \\
\textbf{Class} & \textbf{Class} & \textbf{Original} & \textbf{Balanced} & \textbf{Set} & \textbf{Set} \\
\midrule
NF & FQ & 182,880 & 11,073 & 92,716 & 95,770 \\
NF & A  & 19 & 6 & 44 & 0 \\
NF & B  & 12,130 & 3,639 & 6,210 & 4,472 \\
NF & C  & 18,060 & 5,418 & 8,191 & 10,460 \\
\midrule
Total & NF & 213,089 & \textbf{20,136} & 107,161 & 110,702 \\
\midrule
FL & M  & 3,168 & 19,008 & 1,384 & 1,853 \\
FL & X  & 188 & 1,128 & 154 & 320 \\
\midrule
Total & FL & 3,356 & \textbf{20,136} & 1,538 & 2,173 \\
\bottomrule
\end{tabular}
\end{center}
\label{table:dataset}
\end{table}
We adopt the time-segmented tri-monthly data partitioning strategy originally introduced in~\cite{pandey2021bigdata} and subsequently adapted for localized active region (AR) patches in~\cite{pandeyecml2024}. In contrast to the full-disk approach, which partitions data based on entire solar disk observations, the AR-based scheme performs partitioning at the level of individual active regions, ensuring that each AR is uniquely assigned to a single data split. Specifically, we combine partition-1 and partition-2 as the training set, use partition-3 as the validation set, and reserve partition-4 as the test set. Since flares of class $\geq$M are relatively rare, the dataset exhibits a class imbalance problem. To address this, we apply five domain-relevant data augmentation techniques to the FL-class instances in the training set: (i) vertical flipping, (ii) horizontal flipping, (iii) addition of random noise (up to $\pm$25 G), (iv) Gaussian blurring, and (v) polarity inversion, which involves multiplying all raster values by $-1$ to reverse magnetic polarity. Data augmentation alone does not fully resolve the imbalance. Therefore, we apply random undersampling to the NF-class in the training set, retaining approximately 6 percent of flare-quiet instances and 30 percent of A-, B-, and C-class instances. The overall distribution of binary-labeled AR patch data, including the NF and FL classes along with their respective sub-classes, is summarized in Table~\ref{table:dataset}. For validation and test sets, we preserve the original imbalanced distribution to support a realistic evaluation scenario.

\begin{figure*}[htbp]
\centering
\includegraphics[width=0.95\linewidth ]{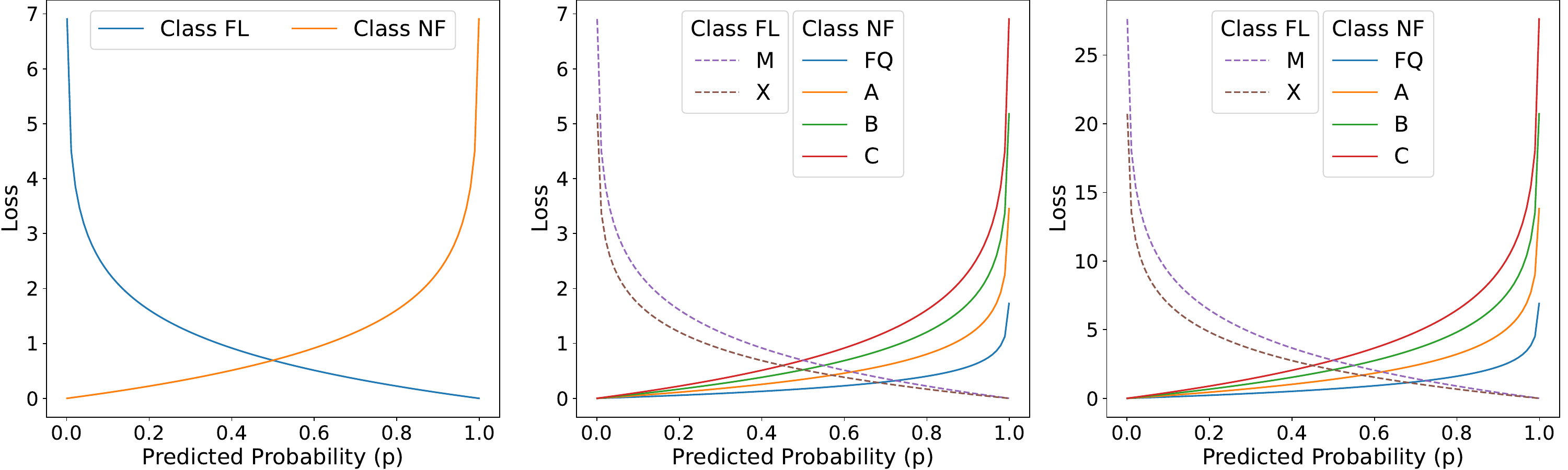}
\caption[]{An illustrative plot depicting: (a) the standard binary cross-entropy (BCE) loss; and (b–c) the BCE with proximity penalty (BCE-PP) used for solar flare prediction, which incorporates ordinal flare characteristics through a loss-weighting mechanism with $\alpha = 0.25$ and $\alpha = 1$, respectively. Note: the FL class corresponds to target~1, and the NF class corresponds to target~0.}
\label{fig:loss}
\end{figure*}

The solar flare prediction task is formulated in this work as a binary image classification problem; accordingly, we employ a lightweight convolutional neural network (CNN) architecture, MobileNet~\cite{mobilenet-v3}. Although attention-based models such as Vision Transformers (ViTs)~\cite{vit} have recently demonstrated state-of-the-art performance in image classification, they typically involve a large number of trainable parameters ($\sim$86--632 million), making them computationally demanding and less suitable for applications with limited resources or relatively small datasets. Therefore, given the modest size of our dataset, we adopt a lightweight model architecture for this study.

\section{Proximity-Penalized Binary Loss (BCE-PP)} \label{sec:method}

In this paper, we build upon our prior work, where we introduced a loss function (BCE-SF; see \cite{dsaa2024} for details) for binary solar flare prediction that incorporates the ordinal characteristics of flare sub-classes into the standard binary cross-entropy (BCE) loss. In the present work, we use this ordinal information as a form of regularization. Formally, let \( N \) denote the total number of instances in a batch. For each sample \( i \), let \( y_i \) represent the true label, where \( y_i \in \{0, 1\} \). The predicted probability that the \( i \)-th sample belongs to the ``FL'' class (target 1) is given by \( p_i = \sigma(\hat{y}_i) \), where \( \hat{y}_i \) is the model output (logit) and \( \sigma \) denotes the sigmoid function. The standard binary cross-entropy loss, \( \text{BCE}(y, \hat{y}) \), is expressed as in Eq.~(\ref{eq:bce}), and the corresponding loss curve is illustrated in Fig.~\ref{fig:loss} (a).

\begin{equation}\label{eq:bce}
    \text{BCE}(y, \hat{y}) = -\frac{1}{N} \sum_{i=1}^N \left[ y_i \log(p_i) + (1 - y_i) \log(1 - p_i) \right]
\end{equation}

As discussed earlier in Sec.~\ref{sec:intro}, under the binary formulation of solar flare prediction with a threshold of $\geq$M, the data are divided into two classes: (i) the NF class, comprising FQ-, A-, B-, and C-class instances, and (ii) the FL class, comprising M- and X-class instances. These flare sub-classes exhibit an inherent ordinal structure, represented as FQ~$<$~A~$<$~B~$<$~C~$<$~M~$<$~X. With the chosen threshold of $\geq$M, FQ and X-class flares become the two extremes from the classification threshold, while C- and M-class flares are at the closest proximity. The traditional BCE loss overlooks this ordinal nature during optimization, treating all incorrect predictions equally. Therefore, we introduce a weighting mechanism based on the ordinal level of each sub-class (flare class within binary class) such that instances closer to the classification threshold are assigned higher weights. This ensures that incorrect predictions of such instances are penalized more compared to the instances closer to the extremes. The weights ($\beta_i$) for an input instance $i$, belonging to a subclass $c_i \in$ \{FQ, A, B, C, M, X\} based on proximity to the threshold and ordinality of the data, can be defined as shown in Eq.~(\ref{eq:beta}).
\begin{equation} \label{eq:beta}
{\beta}_i = \begin{cases}
10 & \text{if } c_i = FQ \\
10^2 & \text{if } c_i = A \\
10^3 & \text{if } c_i = B \text{ or } c_i = X \\
10^4 & \text{if } c_i = C \text{ or } c_i = M \\
\end{cases}
\end{equation}

We utilize these ordinal weights ($\beta_i$) representing individual flare classes, and our proposed binary cross-entropy loss with penalty based on proximity to the prediction threshold (BCE-PP) can be represented as shown in Eq.~(\ref{eq:bcesf}) \footnote{\textit{Note.} Flare classes are inherently defined on a logarithmic scale, with each successive class representing an order-of-magnitude change in the underlying measurement scale. To reflect this, we assign $\beta_i$ values in powers of ten, consistent with the conventional logarithmic spacing between classes. In the loss formulation (Eq.~\ref{eq:bcesf}), we apply $\log_{10}(\beta_i)$, which effectively maps these powers of ten to a simple linear progression (e.g., 1, 2, 3, 4). This transformation retains the ordinal relationship implied by the original logarithmic definition of the classes, but produces a linearly spaced weight set that is more stable for optimization and easier to interpret in the context of the BCE loss.
}.
\begin{equation}\label{eq:bcesf}
    \text{BCE-PP}(y, \hat{y}) = -\frac{1}{N} \sum_{i=1}^N \alpha \times  \text{BCE}(y_i,  \hat{y}_i) \times \log_{10}(\beta_i)
\end{equation}
Here, $\alpha$ serves as a scaling factor that adjusts the magnitude of the loss to align with the corresponding BCE loss scale. When $\alpha = 0.25$, the maximum loss value for an incorrectly predicted instance corresponds to the scale of the BCE loss, as illustrated in Fig.~\ref{fig:loss} (b). In this configuration, the loss values for the C- and M-class instances, those closest to the prediction threshold are comparable to the BCE loss scale, whereas all other incorrect predictions yield smaller loss magnitudes. Conversely, when $\alpha = 1$, the minimum loss value for an incorrectly predicted instance matches the BCE loss scale, as shown in Fig.~\ref{fig:loss} (c). In this case, the FQ instances exhibit loss values similar to those from the BCE loss, while other misclassified instances have proportionally higher losses. Accordingly, we recommend setting $\alpha \in [0.25, 1]$, treating it as a tunable hyperparameter for optimal performance. Notably, the BCE-PP loss leverages the inherent ordinal properties of flare classes to provide a simple yet effective data-driven extension of the BCE loss, without adding model-dependent parameters or computational overhead.

\section{Experimental Evaluation} \label{sec:expt}
\subsection{Experimental Settings}
\begin{table}[htbp]
\setlength{\tabcolsep}{5.5pt}
\renewcommand{\arraystretch}{1.5}

\caption{Hyperparameters search space with optimal hyperparamters for models trained with BCE and BCE-PP loss.}
\begin{center}
 \begin{tabular}{r r r r}
\hline
\multicolumn{2}{r}{}              
&                                            
\multicolumn{2}{c}{Optimal Parameters}\\
Hyperparameters & Search Space & BCE  & BCE-PP \\
\hline
Initial Learning Rate &  \{0.00001 to 0.01\}   &  0.01  & 0.001 \\

Weight Decay &  \{0.00001 to 0.01\}  &  0.01 & 0.001 \\

Batch Size & \{48, 64, 80\} & 64 & 64 \\ 

Scaling Factor ($\alpha$)& \{0.25, 0.5, 0.75, 1\} & N/A & 0.75 \\ 
\hline
\end{tabular}
\end{center}
\label{table:params}
\end{table}
In our hyperparameter selection procedure, we define a search space that includes the initial learning rate, weight decay coefficient, batch size, and scaling factor ($\alpha$), as summarized in Table~\ref{table:params}. We then perform a grid search over this space, evaluating model performance on the validation set for all three architectures. Each model is trained using stochastic gradient descent (SGD) with both BCE and BCE-PP loss functions. To adaptively control the learning rate, we apply the dynamic scheduling strategy \texttt{ReduceLRonPlateau}, using a reduction factor of 0.9 and a patience value of two epochs. This scheduler begins training with the initial learning rate specified in Table~\ref{table:params} and updates it as: 
\(
\texttt{new learning rate} = \texttt{current learning rate} \times \texttt{factor}.
\)
If the validation loss fails to improve for two consecutive epochs, the learning rate is reduced by the specified factor. After completing the grid search and validation, the optimal hyperparameters, reported in Table~\ref{table:params}, were selected based on superior validation performance. These settings were subsequently used to train the final models for 50 epochs before evaluation.

\subsection{Evaluation Metrics}

The True Skill Statistic (TSS; Eq.~\ref{eq:TSS}) and the Heidke Skill Score (HSS; Eq.~\ref{eq:HSS}), both computed from the four components of the confusion matrix including true positives (TP), true negatives (TN), false positives (FP), and false negatives (FN), are widely used performance metrics for evaluating solar flare prediction models~\cite{Ahmadzadeh2019, Ahmadzadeh2021}. In the context of this paper, FL indicates positive class and NF indicates negative class.

\begin{equation}\label{eq:TSS}
    TSS = \frac{TP}{TP+FN} - \frac{FP}{FP+TN} 
\end{equation}
\begin{equation}\label{eq:HSS}
    HSS = 2\times \frac{TP \times TN - FN \times FP}{((P \times (FN + TN) + (TP + FP) \times N))}
\end{equation}
\begin{center}
where, $N = TN + FP$ and  $P = TP + FN$.    
\end{center}

TSS and HSS values range from $-1$ to $1$, where a value of $1$ indicates perfect prediction, $-1$ corresponds to all predictions being incorrect (equivalently, all inverse predictions being correct, implying skill), and $0$ denotes no predictive skill. Unlike TSS, HSS accounts for class imbalance and is therefore commonly used for evaluating solar flare prediction models, given the pronounced imbalance present in flare datasets~\cite{Ahmadzadeh2019, Ahmadzadeh2021}. However, selecting a candidate model based solely on these two metrics can be challenging, as it requires prioritizing one measure over the other. To address this, we combine TSS and HSS through their geometric mean to define a Composite Skill Score (CSS; Eq.~\ref{eq:CSS}), providing a unified metric that balances discriminative power and imbalance sensitivity~\cite{pandeyecml2024, dsaa2024}. Accordingly, we use CSS as the primary evaluation metric while also reporting TSS and HSS for completeness.

\begin{equation}\label{eq:CSS}
CSS = 
\begin{cases}
0, & \text{if } TSS <0 \quad \text{or} \quad HSS < 0 \\
\sqrt{TSS \times HSS}, & \text{otherwise}
\end{cases}
\end{equation}


\subsection{Evaluation}
\begin{table}[tbh!]
\vspace{-10pt}
\setlength{\tabcolsep}{5 pt}
\renewcommand{\arraystretch}{1.5}
\caption{The performance evaluation of models trained with BCE and BCE-PP on both validation and test set.}
\vspace{-10pt}
\begin{center}
\begin{tabular}{rrrrrrrr}
\toprule
\multicolumn{8}{c}{\textbf{Performance Evaluation on Validation Set}}\\
Models & TP & FP & TN & FN & TSS & HSS & CSS \\
\midrule
BCE &1,057 & 5,102 & 102,059 & 481& 0.64 & 0.26 & 0.41 \\
\textbf{BCE-PP} &973 & \textbf{2,969} & 104,192 & 565 & 0.61 & 0.34 & \textbf{0.45} \\
\midrule
\multicolumn{8}{c}{\textbf{Performance Evaluation on Test Set}}\\
Model & TP & FP & TN & FN & TSS & HSS & CSS \\
\midrule
BCE &1,548 & 5,429 & 105,273 & 625 & 0.66 & 0.31 & 0.45 \\
\textbf{BCE-PP} &1,446 & \textbf{4,460} & 106,242 & 727 & 0.63 & 0.34 & \textbf{0.46} \\
\bottomrule
\end{tabular}
\end{center}
\label{table:validation_results}
\vspace{-10pt}
\end{table}

As mentioned earlier in Sec.~\ref{sec:data}, we utilized ``train-val-test'' split of the dataset in our experiments. We observed that the model trained with standard BCE loss achieves a CSS$\sim$0.41 (TSS$\sim$0.64 and HSS$\sim$0.26) and CSS$\sim$0.45 (TSS$\sim$0.66 and HSS$\sim$0.31) on validation and test set respectively. Upon comparing this with the performance of proposed BCE-PP loss, we observed an improvement of $\sim$4\% and $\sim$1\% in terms of CSS on validation and test set respectively. This demonstrates that the BCE-PP loss leads to better performance compared to the standard BCE loss. Furthermore, it is important to note that, while TSS scores are high for BCE-trained model in both evaluation set, HSS scores are consistently low compared to BCE-PP trained model. Therefore, a single metric like CSS, as discussed, can be effective during model selection. The detailed results along with the confusion matrices are shown in Table.~\ref{table:validation_results}.

Moreover, we observe that the models optimized with BCE-PP generate significantly fewer false positives (FP) compared to the model trained with BCE, although both deliver similar performance on the test set in terms of CSS. Furthermore, the improvement in terms of FPs delivered by BCE-PP comes with slightly higher false negatives (FNs) compared to the BCE loss. This increment in FN counts may be due to higher class imbalance in both the validation and test sets and shows a trade-off between FP and FN, while the performance in terms of skill score might be comparable.

\begin{table}[tbh!]
\setlength{\tabcolsep}{5 pt}
\renewcommand{\arraystretch}{1.5}

\caption{Performance comparison with prior work in terms of TSS, HSS, and CSS.}
\vspace{-10pt}
\begin{center}
 \begin{tabular}{l l c c c}
\hline
\multicolumn{2}{l}{}              
&                                            
\multicolumn{3}{c}{Evaluation Metrics} \\
Model & Backbone & TSS  & HSS & CSS \\
\hline

 Pandey et al., 2024(a) \cite{pandeyecml2024} & MobileNet & 0.59 & \textbf{0.44} & \textbf{0.51} \\

 Pandey et al., 2024(b) \cite{dsaa2024} & ResNet (BCE-SF)& 0.58 & 0.38 & 0.47 \\

This Work & MobileNet (BCE-PP) & \textbf{0.63} & 0.34 & 0.46 \\

\hline
\end{tabular}
\end{center}
\label{table:performance}
\end{table}

We compare our proposed BCE-PP model against two previously published approaches for binary solar flare prediction, using TSS, HSS, and CSS as evaluation metrics (Table~\ref{table:performance}). While the model trained with BCE-PP achieves the highest TSS (0.63), it falls short in terms of CSS and HSS compared to the class-weighted BCE model from \cite{pandeyecml2024}, which shows the best overall CSS (0.51). The ordinal-encoded BCE-SF variant from \cite{dsaa2024} underperforms across all metrics. One possible reason for the lower CSS and HSS in our approach is the reduced training data coverage: both prior models are trained on flare events across the full $\pm90^\circ$ solar disk (using a larger number of training instances), whereas our BCE-PP based model is trained on a subset limited to $\pm60^\circ$. All models are validated and tested on the same data partitions, making training data volume a potential factor contributing to performance differences. While BCE-PP does not outperform across all metrics, its design introduces a mechanism for handling boundary ambiguity, which may offer utility when applied under more balanced training regimes.

\section{Conclusion and Future Work} \label{sec:conc}
In this study, we proposed an ordinal boundary-aware binary loss function to optimize data-driven models for solar flare prediction. By encoding ordinal relationships among flare sub-classes, the loss introduces a soft margin around the decision threshold, which helps reduce false positives and improves model regularization. Although our results show only marginal improvements over standard binary loss formulations, particularly under constrained data settings (limited to $\pm$60$^\circ$ solar longitude), this work offers a new direction for incorporating ordinal structure into binary classification. Future extensions may include leveraging actual peak X-ray flux values as continuous ordinal targets, integrating multimodal solar observations, exploring spatiotemporal architectures, and incorporating interpretability to improve model trust and robustness. 
\bibliographystyle{IEEEtran}
\bibliography{references}

\end{document}